\documentclass[conference]{IEEEtran}
\IEEEoverridecommandlockouts
\usepackage{cite}
\usepackage{amsmath,amssymb,amsfonts}
\usepackage{algorithmic}
\usepackage{graphicx}
\usepackage{amsthm}
\usepackage{textcomp}
\usepackage{xcolor}
\def\BibTeX{{\rm B\kern-.05em{\sc i\kern-.025em b}\kern-.08em
    T\kern-.1667em\lower.7ex\hbox{E}\kern-.125emX}}
    
\newcommand{\sig}[1]{{\small\textsf{{#1}}}}    
\newtheorem{defi}{Definition}
\newtheorem{lemma}{Lemma}
    
\usepackage{tikz}
\usepackage{textcomp}
\usepackage{hyperref}
\newcommand\copyrighttext{%
  \footnotesize \textcopyright 2021 IEEE. Personal use of this material is permitted.
  Permission from IEEE must be obtained for all other uses, in any current or future
  media, including reprinting/republishing this material for advertising or promotional
  purposes, creating new collective works, for resale or redistribution to servers or
  lists, or reuse of any copyrighted component of this work in other works.
  DOI: 10.23919/DATE51398.2021.9473957}
\newcommand\copyrightnotice{%
\begin{tikzpicture}[remember picture,overlay]
\node[anchor=south,yshift=10pt] at (current page.south) {\fbox{\parbox{\dimexpr\textwidth-\fboxsep-\fboxrule\relax}{\copyrighttext}}};
\end{tikzpicture}%
}

\begin{document}

\title{Provably-Robust Runtime Monitoring of Neuron Activation Patterns}


\author{\IEEEauthorblockN{Chih-Hong Cheng}\thanks{This research is part of FOCETA project that has received funding from the European Union’s Horizon 2020 research and innovation programme under grant agreement No 956123.}
\IEEEauthorblockA{
\textit{DENSO AUTOMOTIVE Deutschland GmbH}\\
Email: c.cheng@eu.denso.com}
}

\maketitle
\copyrightnotice

\begin{abstract}

For deep neural networks (DNNs) to be used in safety-critical autonomous driving tasks, it is desirable to monitor in operation time if the input for the DNN is similar to the data used in DNN training. While recent results in monitoring DNN activation patterns provide a sound guarantee due to building an abstraction out of the training data set, reducing false positives due to slight input perturbation has been an issue towards successfully adapting the techniques. We address this challenge by integrating formal symbolic reasoning inside the monitor construction process. The algorithm performs a sound worst-case estimate of neuron values with inputs (or features) subject to perturbation, before the abstraction function is applied to build the monitor. The provable robustness  is further generalized to cases where monitoring a single neuron can use more than one bit, implying that one can record activation patterns with a fine-grained decision on the neuron value interval.

\end{abstract}


\section{Introduction}

For autonomous driving, perception modules implemented using deep neural networks (DNNs) are trained using data collected within the \emph{operational design domain} (ODD). The generalization capability of machine learning implicitly implies that a well-trained DNN can only perform well on inputs \emph{similar} to the those collected in the training data set.\footnote{In layman words, DNNs are good at interpolation but bad at extrapolation.} As a consequence, for autonomous vehicles operating in an \emph{open environment} where ``black swans'' undoubtedly  occurs, one shall design proper monitoring mechanisms to detect if the perception system encounters any input \emph{distant} from the training data. Upon detection,  perform appropriate actions such as functional degradation or conservative maneuvers. 

In this paper, we consider monitors based on building an \emph{abstraction out of  neuron activation patterns} of the training data~\cite{cheng2019runtime,henzinger2019outside}. As for inputs being pixels of images, it is impossible to create a compact set representation directly on the input values, the underlying idea is to select a set of neurons to be monitored, where the monitored neurons within close-to-output layers represent high-level features. For each input to the DNN, values of all monitored neurons form a feature vector. Monitor construction is to algorithmically build a compact set representation that guarantees to contain all feature vectors (via methods such as Boolean abstraction or abstraction by \textsf{min} and \textsf{max} operators). Due to abstraction, the monitor provides a \emph{sound} guarantee: A warning over an input implies that no input in the training data set is ``close'' with the transformed feature defining the closeness. The soundness of the monitor offers a clear argument for understandability, certifiably, and homologation purposes. 

While previous results in activation-based monitors have demonstrated their capability in well-engineered data sets such as MNIST or GTSRB, our deployment of the monitor in a physical laboratory setting (race track) reveals unexpected challenges. In the real-world environment, the training data is only finitely sampled from the ODD, and there is potential intrinsic randomness of the entire data collection process subject to factors such as tiny changes of lighting conditions in the day. Consequently, although the training data set is sufficient to build a perception unit, monitoring construction methods that enlarge the abstraction using the validation data set are insufficient to cure aleatory uncertainty.
In other words, the monitor can generate many false alarms (i.e., the vehicle is in the ODD and nothing strange occurs, but the monitor signals a warning) to an unaccepted level (e.g., $1\%$) that is inappropriate to be used in autonomous driving. 

Towards the above issues, we design a new monitor construction algorithm with \emph{robustness guarantees}. The underlying idea is to replace each feature vector $f(v_{tr})$ generated by an input $v_{tr}$ propagating through the DNN with a conservative bound~$[\textsf{min}_{\delta \in \Delta}(f(v_{tr} +\delta)), \textsf{max}_{\delta \in \Delta}(f(v_{tr} +\delta))]$ which considers all $\Delta$-bounded perturbation over~$v_{tr}$. The abstraction (originally applied on the set of feature vectors) is now applied on the set of conservative bounds. The bound can be computed using state-of-the-art DNN verification methods such as boxed abstraction (interval bound propagation)~\cite{gowal2018effectiveness}, 
zonotope abstraction~\cite{gehr2018ai2}, or star sets~\cite{tran2019star}. The robustness implies that one can formally guarantee that any data point being close in the input domain (subject to $\Delta$) can never trigger a violation in the monitor whose ``closeness'' is defined in the feature vector space. Finally, we also extend the robust monitor concept to a generalized monitor setup where monitoring a single neuron can use more than one bit. The multi-bit setup implies that it can record activation patterns with a fine-grained decision on the neuron value interval.

The rest of the paper is structured as follows. Section~\ref{sec.related} highlights the related work and provides a general comparison over other approaches. Section~\ref{sec.monitor} presents the basic formulation of DNN, algorithms for constructing provably robust monitors, and extensions to monitors beyond single-bit on-off activation. Finally in Section~\ref{sec.conclusion}, we summarize our findings in the experiment and conclude with future directions.

\section{Related Work}\label{sec.related}

The problem of understanding uncertainty in the decision making of a neural network has been extensively studied. Bayesian approaches~\cite{blundell2015weight,hernandez2015probabilistic} learn a probability distribution on the weights of a neural network. Monte-Carlo approaches use an ensemble of networks to estimate the uncertainty; the ensemble can be generated either by differently trained networks~\cite{lakshminarayanan2017simple} or by performing drop-out in operation~\cite{gal2016dropout}. The ensemble nature of Monte-Carlo approaches implies that it is costly to execute. The complementing work on monitoring neuron activation patterns~\cite{cheng2019runtime,henzinger2019outside,cheng2020towards,lukina2020into} can derive formal guarantees as it uses sound abstraction over the training data set. Both qualitative decisions~\cite{cheng2019runtime,henzinger2019outside} and quantitative decisions~\cite{lukina2020into} are made possible, and it is also used in assume-guarantee based reasoning for safety verification of highway autonomous driving systems~\cite{cheng2020towards}. Our work complements current results in abstraction-based neuron monitoring and can be integrated into these work with ease; it adds a new dimension of integrating symbolic reasoning inside the monitor construction process, thereby providing robustness guarantees. 

Finally, the technique used in this paper is highly related to the work on  verification of neural networks using abstract interpretation (e.g.,~\cite{gowal2018effectiveness,gehr2018ai2,tran2019star}), where the symbolic reasoning is used not in the engineering of neural networks but rather in the construction of monitors to be used in operation.

\section{Robust Neuron Activation Pattern Monitors}\label{sec.monitor}

\subsection{DNN and Runtime Monitors}

A feed-forward DNN model consists of an input layer, multiple hidden layers and an output layer. We consider only DNNs after training, i.e., all model parameters (e.g., weights, bias) are fixed. With fixed parameters, a DNN model is a function $\mathcal{G}: \mathbb{R}^{d_0}\rightarrow \mathbb{R}^{d_n}$, where~$d_0$ is the input dimension and~$d_{n}$ is the output dimension. The model~$\mathcal{G}$ is built out of a sequence of functions $\mathcal{G} :=g^{n} \otimes g^{n-1} \otimes \ldots \otimes  g^2 \otimes
g^1$, where $n$ is the number of layers in the DNN, and for $k \in \{1, \ldots, n\}$, $g^k : \mathbb{R}^{d_{k-1}} \rightarrow \mathbb{R}^{d_{k}}$ is the transformation in the $k$-th layer. Given an input $v_{tr} \in \mathbb{R}^{d_0}$, the computation of $\mathcal{G}$  is $\mathcal{G}^n(v_{tr})=g^n(g^{n-1}(\ldots (g^2 (g^1(v_{tr})))\ldots))$, i.e., to perform functional composition over $g^1, \ldots, g^n$. 

For the ease of explanation, here we only formulate the simplistic monitoring setup: the abstraction-based monitor applies to a specific layer, all neurons in the layer are monitored, the decision is qualitative, and there is no additional abstraction enlargement by using the validation set. Extensions such as configuring to multi-layer monitoring as well as selecting a subset of neurons to be monitored are straightforward. We use $\mathcal{G}^k$ to abbreviate $g^{k} \otimes g^{k-1} \otimes \ldots \otimes  g^2 \otimes g^1$, and use $\mathcal{G}^{l\hookrightarrow k}$ to abbreviate $g^{k} \otimes g^{k-1} \otimes \ldots \otimes  g^{l+1} \otimes g^{l}$. For the sake of simplicity in the established theorem, we abuse the notation and may use $\mathcal{G}^0(v_{tr})$ to represent~$v_{tr}$ (i.e., no transformation applied). 
Lastly, for vector~$v$ and function $f$ that outputs a vector, we use subscripts $v_j$ and $f_j$ to indicate the $j$-th value.

Let $\mathcal{D}_{tr} \subset \mathbb{R}^{d_0}$ be the training data set, then a generic algorithm for building a \emph{neuron activation pattern monitor} $\mathcal{M}^{\langle \mathcal{G}, k \rangle}$ is as follows:

\vspace{1mm}

\begin{algorithmic}[h]
\STATE $\mathcal{M}^{\langle \mathcal{G}, k \rangle}\gets \mathcal{M}_{0}$
 \FOR{$v_{tr} \in \mathcal{D}_{tr}$} 
    \STATE {$\mathcal{M}^{\langle \mathcal{G}, k \rangle} \gets\mathcal{M}^{\langle \mathcal{G}, k \rangle} \uplus (\sig{ab}(\mathcal{G}^{k}(v_{tr})))$} 
\ENDFOR
\end{algorithmic}

Different abstraction techniques may be applied, by setting different initial value $\mathcal{M}_{0}$, operator~$\uplus$ and function~$\sig{ab}$. 

\begin{itemize} 
    \item For monitoring the minimum and maximum  values~\cite{cheng2020towards,henzinger2019outside}, $\mathcal{M}^{\langle \mathcal{G}, k \rangle}$  equals $\langle (L_1, U_1),  \ldots,  (L_{d_k}, U_{d_k})\rangle$ where for $j \in \{1, \ldots, d_k\}$, the pair $(L_j, U_j)$ tracks for the complete training data set, the minimum and the maximum value of the $j$-th neuron. Precisely,
        \begin{itemize}
            \item $\mathcal{M}_0$  equals $\langle (\infty, -\infty), \ldots, (\infty, -\infty)\rangle$,
           
            \item  $\sig{ab}$ is the identity function, and 
            \item $\mathcal{M}^{\langle \mathcal{G}, k \rangle} \uplus (v_1, \ldots v_{d_k}) := \langle (\sig{min}(L_1, v_1),\\ \sig{max}(U_1, v_1)),  \ldots, (\sig{min}(L_{d_k}, v_{d_k}), \sig{max}(U_{d_k}, v_{d_k}))\rangle$. 
        \end{itemize}
    
    In operation, given an input $v_{op} \in \mathbb{R}^{d_0}$, the monitor~$\mathcal{M}^{\langle \mathcal{G}, k \rangle}$ \emph{raises a warning}, denoted as $\mathcal{M}^{\langle \mathcal{G}, k \rangle}(v_{op}) = \sig{true}$, if $\exists j \in \{1, \ldots, d_k\}$ such that $\mathcal{G}^{k}_j(v_{op}) < L_j$ or $\mathcal{G}^{k}_j(v_{op}) > U_j$. 
    
    \vspace{1mm}
    \item For monitoring on-off activation patterns introduced in~\cite{cheng2019runtime}, the construction performs  Boolean abstraction based on a pre-defined threshold (e.g., sign of the neuron value, or average of all visited values) to build a Boolean word. The set of Boolean words can be efficiently stored using \emph{Binary Decision Diagrams (BDDs)}~\cite{bryant1992symbolic}. 
            \begin{itemize}
            \item $\mathcal{M}_0 = \emptyset$.
            \item Let $c_1, c_2, \ldots, c_{d_{k}}$ be predefined constants. The function $\sig{ab} (v_1, \ldots v_{d_k})$ returns $(b_1, \ldots, b_{d_k})$ where for $j \in \{1, \ldots, d_{k}\}$, 
            \begin{itemize}
                \item $b_j = \sig{1}$ if $v_j > c_j$, and
                \item $b_j = \sig{0}$ otherwise.
            \end{itemize}
            
            \item $\mathcal{M}^{\langle \mathcal{G}, k \rangle} \uplus (b_1, \ldots, b_{d_k}) := \mathcal{M}^{\langle \mathcal{G}, k \rangle}\cup \{(b_1, \ldots, b_{d_k})\}$. 
        \end{itemize}
        Given an input $v_{op} \in \mathbb{R}^{d_0}$, the monitor raises a warning, denoted as $\mathcal{M}^{\langle \mathcal{G}, k \rangle}(v_{op}) = \sig{true}$, if $\sig{ab}(\mathcal{G}^{k}(v_{op})) \not\in \mathcal{M}$. 
\end{itemize}

\subsection{Robust Monitors}

To build robust monitors, we first define feature-level perturbation to capture the effect on the monitored layer via a worst-case estimate. 

\vspace{1mm}
\begin{defi}
Given $v_{tr} \in \mathbb{R}^{d_0}$ and an integer $k_p$ where $0 \leq k_p < k$, define the \emph{perturbation estimate} $\sig{pe}^{\mathcal{G}}_k(v_{tr}, k_p, \Delta) = \langle (l_1, u_1), \ldots, (l_{d_k}, u_{d_k})\rangle$ which guarantees the following property: For $\delta_1, \ldots, \delta_{k_p} \in\mathbb{R}$ where $|\delta_1|, \ldots, |\delta_{k_p}| \leq \Delta$, let $\breve{v} := (\breve{v}_{1},\ldots, \breve{v}_{k_p})\in \mathbb{R}^{d_{k_p}}$ where  $\forall j \in \{1, \ldots, d_k\}: \mathcal{G}^{k_p}_{j}(v_{tr}) + \delta_j = \breve{v}_{j}$. Then the following condition holds.

\begin{equation}
    \forall j \in \{1, \ldots, d_k\}: l_j \leq \mathcal{G}^{{k_p}+1 \hookrightarrow k}_j(\breve{v}) \leq u_j
\end{equation}

\end{defi}

Intuitively, the estimate considers perturbation occurring at the output of layer~$k_p$, with the amount of perturbation in each dimension being bounded by~$\Delta$. When $k_p = 0$, the perturbation appears at the input layer. Any perturbed vector~$\breve{v}$, after passing through the sub-network  $g^{k} \otimes g^{k-1} \otimes \ldots \otimes  g^{k_p+1}$, shall produce a vector $\mathcal{G}^{{k_p}+1 \hookrightarrow k}(\breve{v})$ with every dimension being bounded by 
$[l_j, u_j]$ as provided by the perturbation estimate. The perturbation estimate can be efficiently computed using boxed abstraction (interval bound propagation)~\cite{gowal2018effectiveness}, 
zonotope abstraction~\cite{gehr2018ai2}, or star sets~\cite{tran2019star}. In our implementation, boxed abstraction is used. 

Let $\mathcal{D}_{tr} \subset \mathbb{R}^{d_0}$ be the training data set, then a generic algorithm for building a \emph{robust neuron activation pattern monitor} $\mathcal{M}^{\langle \mathcal{G}, k, k_p, \Delta \rangle}$ is as follows:

\vspace{1mm}

\begin{algorithmic}[h]
\STATE $\mathcal{M}^{\langle \mathcal{G}, k, k_p, \Delta \rangle}\gets \mathcal{M}_{0}$
 \FOR{$v_{tr} \in \mathcal{D}_{tr}$} 
    \STATE {$\mathcal{M}^{\langle \mathcal{G}, k, k_p, \Delta \rangle} \gets \mathcal{M}^{\langle \mathcal{G}, k, k_p, \Delta \rangle} \uplus_{R} \sig{ab}_{R}(\sig{pe}^{\mathcal{G}}_k(v_{tr}, k_p, \Delta))$} 
\ENDFOR
\end{algorithmic}

We again consider previously stated two types of monitors (min-max  and Boolean abstraction), by detailing the corresponding initial values $\mathcal{M}_{0}$, operator~$\uplus_{R}$ and function~$\sig{ab}_{R}$ that have different operations due to inputs being bounds. 

\begin{itemize} 
    \item For robust monitoring the minimum and maximum visited neuron values, $\mathcal{M}^{\langle \mathcal{G}, k, k_p, \Delta \rangle}$  equals $\langle (L_1, U_1),  \ldots,  (L_{d_k}, U_{d_k})\rangle$.
        \begin{itemize}
            \item $\mathcal{M}_0$  equals $\langle (\infty, -\infty), \ldots, (\infty, -\infty)\rangle$,
           
            \item  $\sig{ab}_{R}$ is the identity function, and 
            \item $\mathcal{M}^{\langle \mathcal{G}, k, k_p, \Delta \rangle} \uplus_{R} \langle (l_1, u_1), \ldots, (l_{d_k}, u_{d_k})\rangle := \\  \langle (\sig{min}(L_1, l_1),  \sig{max}(U_1, u_1)), \ldots, (\sig{min}(L_{d_k}, l_{d_k}), \\ \sig{max}(U_{d_k}, u_{d_k}))\rangle$. 
        \end{itemize}
    
    In other words, the monitor considers the worst case estimate (under perturbation) and enlarges the bound correspondingly. 
    
    \vspace{1mm}
    \item For robust monitoring on-off activation patterns,
            \begin{itemize}
            \item $\mathcal{M}_0 = \emptyset$.
            \item Let $c_1, c_2, \ldots, c_{d_{k}}$ be predefined constants. The function $\sig{ab}_{R} (\langle (l_1, u_1), \ldots, (l_{d_k}, u_{d_k})\rangle)$ returns $(b_1, \ldots, b_{d_k})$ where for $j \in \{1, \ldots, d_{k}\}$,  
            \begin{itemize}
                \item set $b_j = \sig{1}$  if $l_j > c_j$,
                \item set $b_j = \sig{0}$  if $u_j \leq c_j$,
                \item otherwise, set $b_j$ to ``$ \sig{-}$'', where symbol ``\sig{-}'' means ``don't care'', i.e., the value can be either \sig{1} or \sig{0}.
            \end{itemize}
            \item $\mathcal{M}^{\langle \mathcal{G}, k, k_p, \Delta \rangle} \uplus_{R} (b_1, \ldots, b_{d_k}) :=  \mathcal{M}^{\langle \mathcal{G}, k, k_p, \Delta \rangle} \cup \sig{word2set}(b_1, \ldots, b_{d_k})$, where $\sig{word2set}()$ is the function that returns the set of all binary words by translating ``\sig{-}'' into \sig{0} and \sig{1}. Note that when using BDDs, the translation $\sig{word2set}()$ does not create an exponential blow-up.\footnote{Consider the case where $(b_1, b_2, b_3, b_4) = ( \textsf{1}, \textsf{-}, \textsf{-}, \textsf{0})$. When using BDDs, the set is simply represented as $b_1 \wedge  \neg b_4$, i.e., both $b_3$ and $b_4$ are unconstrained.} 
        \end{itemize}
\end{itemize}

Given an input $v_{op} \in \mathbb{R}^{d_0}$ encountered in operation time, as there is no perturbation (i.e., $\Delta = 0$), it implies that for every $j \in \{1, \ldots, d_k\}$, the computed lower-bound~$l_j$, the upper-bound~$u_j$ and $\mathcal{G}^{k}_{j}(v_{op})$ are equal. One can simply use the condition as stated in the standard monitor to perform abnormality detection. 

\begin{itemize}
    \item For min-max monitors, given an input $v_{op} \in \mathbb{R}^{d_0}$, the monitor~$\mathcal{M}^{\langle \mathcal{G}, k, k_p, \Delta \rangle}$ raises a warning, denoted as $\mathcal{M}^{\langle \mathcal{G}, k, k_p, \Delta \rangle}(v_{op}) = \sig{true}$, if $\exists j \in \{1, \ldots, d_k\}$ such that $\mathcal{G}^{k}_j(v_{op}) < L_j$ or $\mathcal{G}^{k}_j(v_{op}) > U_j$. 
    \item  For Boolean abstraction monitors,    given an input $v_{op} \in \mathbb{R}^{d_0}$, the monitor raises a warning, denoted as $\mathcal{M}^{\langle \mathcal{G}, k, k_p, \Delta \rangle}(v_{op}) = \sig{true}$, if $\sig{ab}(\mathcal{G}^{k}(v_{op})) \not\in \mathcal{M}$. 
\end{itemize}

The abstraction kept by the robust monitor, upon construction, already accounts for all possible cases caused by perturbation bounded by~$\Delta$ at layer~$k_p$. Therefore, for any input similar to the training data, where the similarity is defined using $\Delta$ and $k_p$, the robust monitor can guarantee not to raise a warning. 

\vspace{1mm}
\begin{lemma}
Given a robust monitor $\mathcal{M}^{\langle \mathcal{G}, k, k_p, \Delta \rangle}$, for any input $v_{op} \in \mathbb{R}^{d_0}$, if  $\mathcal{M}^{\langle \mathcal{G}, k, k_p, \Delta \rangle}(v_{op}) = \sig{true}$, then there  \textbf{does not exist} an input $v_{tr} \in \mathcal{D}_{tr}$ s.t. the following condition holds:
\begin{equation}
    \forall j \in \{1, \ldots, d_{k_p}\}: |\mathcal{G}^{k_p}_j(v_{op}) - \mathcal{G}^{k_p}_j(v_{tr}) | \leq \Delta
\end{equation}
\end{lemma}

\subsection{Robust Neuron Interval Activation Monitors}

Finally, we propose a generic setting as an extension of simple min-max and on-off activation monitors. In contrast to only using one bit to monitor a neuron, one can use multiple bits to encode the \emph{interval} where the neuron value is located. For ease of explanation, we use an example of~$2$ bits. For $j \in \{1, \ldots, d_{k}\}$, let $c_{j1}, c_{j2}, c_{j3}$ be predefined constants where $c_{j1} < c_{j2} < c_{j3}$. 

\subsubsection{Standard Interval Activation Monitors}

For standard monitors, we have the following formulation. 
            \begin{itemize}
            \item $\mathcal{M}_0 = \emptyset$.
            \item The function $\sig{ab} (v_1, \ldots v_{d_k})$ returns $(b_1, \ldots, b_{d_k})$ where for $j \in \{1, \ldots, d_{k}\}$, 
            \begin{itemize}
                \item $b_j = \sig{11}$ if $v_j > c_{j3}$, 
                 \item $b_j = \sig{10}$ if $c_{j3} \geq v_j \geq c_{j2}$, 
                 \item $b_j = \sig{01}$ if $c_{j2} > v_j > c_{j1}$, 
                 \item $b_j = \sig{00}$ otherwise.
            \end{itemize}
            In other words, for the interval activation monitor, we allow the monitor to 
            use two bits to encode the value.\footnote{It is a generalization of min-max monitor, as one can set $c_{j3}$ to be the maximum visited neuron value, $c_{j2}$ to be the minimum visited neuron value, and $c_{j1} = -\infty$. It is also a generalization of on-off monitor, by setting $c_{j3} = \infty$ and $c_{j1} = -\infty$. }   
            
            \item $\mathcal{M}^{\langle \mathcal{G}, k \rangle} \uplus (b_1, \ldots b_{d_k}) := \mathcal{M}^{\langle \mathcal{G}, k \rangle}\cup \{(b_1, \ldots, b_{d_k})\}$. 
        \end{itemize}

Given an input $v_{op} \in \mathbb{R}^{d_0}$, the monitor raises a warning, denoted as $\mathcal{M}^{\langle \mathcal{G}, k \rangle}(v_{op}) = \sig{true}$, if $\sig{ab}(\mathcal{G}^{k}(v_{op})) \not\in \mathcal{M}$. 

\vspace{2mm}

\subsubsection{Robust Interval Activation Monitors} 

The following formulation extends the concept to a robust monitoring setting. 
\begin{itemize}
\item $\mathcal{M}_0 = \emptyset$,
\item The function $\sig{ab}_{R} (\langle (l_1, u_1), \ldots, (l_{d_k}, u_{d_k})\rangle)$ returns $(b_1, \ldots, b_{d_k})$ where for $j \in \{1, \ldots, d_{k}\}$,  
\begin{itemize}
                \item $b_j = \{\sig{11}\}$ if $l_j > c_{j3}$,
                 \item $b_j = \{\sig{10}\}$ if $c_{j3} \geq u_j \geq l_j \geq c_{j2}$, 
                 \item $b_j = \{\sig{01}\}$ if $c_{j2}  > u_j \geq l_j > c_{j1}$, 
                 \item $b_j =\{\sig{00}\}$ if $c_{j1}  \geq u_j$,
                 \item $b_j =\{\sig{00, 01}\}$ if $c_{j2}  > u_j > c_{j1}$ amd $c_{j1}  \geq l_j$,
                 \item $b_j = \{\sig{01, 10}\}$ if $c_{j3} \geq u_j  \geq c_{j2}$ and $c_{j2}  >  l_j > c_{j1}$, 
                 \item $b_j = \{\sig{10, 11}\}$ if $u_j > c_{j3}$ and $c_{j3}  \geq l_j \geq c_{j2}$, 
                 \item $b_j =\{\sig{00, 01, 10}\}$ if $c_{j1}  \geq l_j$ and  $c_{j3}  \geq u_j \geq c_{j2}$.
                  \item $b_j = \{\sig{01, 10, 11}\}$ if $u_j > c_{j3}$ and $c_{j2}  > l_j > c_{j1}$, 
                  \item otherwise, $b_j = \{\sig{00, 01, 10, 11}\}$.
\end{itemize}

In contrast to the standard monitor where the value $v_j$ can only fall into one interval, the robust monitor maintains the conservative lower-bound~$l_j$ and upper-bound~$u_j$, and it is possible that~$l_j$ and~$u_j$ do not fall into the same interval, as illustrated in Figure~\ref{fig.interval}. 

\begin{figure}[t]
\centering
\includegraphics[width=0.9\columnwidth]{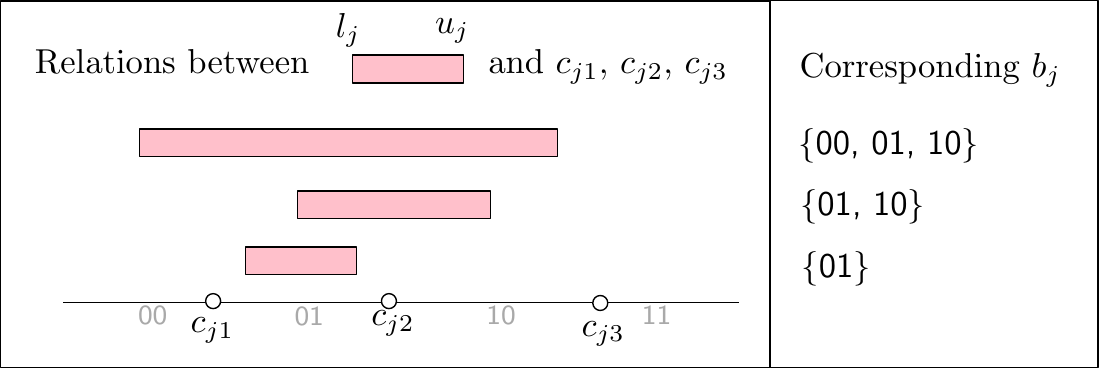}
\vspace{-2mm}
\caption{Understanding the encoding for robust interval activation monitors}
\label{fig.interval}
\vspace{-3mm}
\end{figure}

\item $\mathcal{M}^{\langle \mathcal{G}, k, k_p, \Delta \rangle} \uplus_{R} (b_1, \ldots, b_{d_k}) :=  \mathcal{M}^{\langle \mathcal{G}, k, k_p, \Delta \rangle} \cup \sig{word2set}(b_1, \ldots, b_{d_k})$, where $\sig{word2set}()$ is the function that returns the set of all binary words 
$$\{(\hat{b_1}, \ldots, \hat{b_{d_k}}) \; | \; \forall i \in \{1, \ldots, d_k\}: \hat{b_i} \in b_i \}$$
\end{itemize}

The function $\sig{word2set}$ can be efficiently implemented in BDD, and here we only explain the process using an example where $(b_1, b_2, b_3) = (\{\sig{00, 01}\}, \{\sig{01,10,11}\}, \{\sig{10}\})$. For~$b_1$ (similarly for $b_2$ and $b_3$), one uses two BDD variables $b_{10}, b_{11}$ to encode the first and the second bit. Therefore, the BDD encoding of $\sig{word2set}(b_1, b_2, b_3)$ simply returns $(\neg b_{10})\wedge (b_{20} \vee b_{21})\wedge(b_{30} \wedge \neg b_{31})$.


\begin{figure}[t]
\centering
\includegraphics[width=0.9\columnwidth]{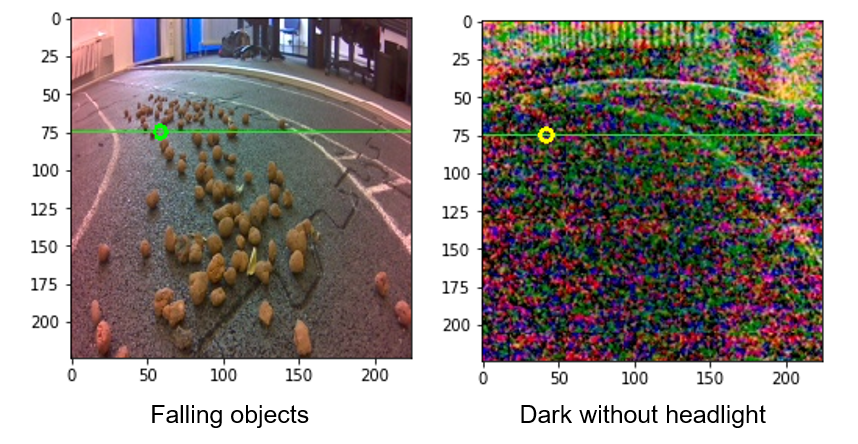}
\vspace{-2mm}
\caption{Experimenting monitoring techniques in the lab setting. }
\label{fig.experiment}
\vspace{-3mm}
\end{figure}

\section{Concluding Remarks}\label{sec.conclusion}

In this paper, we addressed the robustness of activation-pattern monitoring and proposed integrating the symbolic reasoning in the monitor construction process. 
As established in the new monitor construction algorithm, the provable robustness claims ensure that any input data being similar (subject to $k_p$ and $\Delta$) to the training data set will not cause the monitor to raise a warning. 
Therefore, the new monitor reduces false positives and only focuses on inputs that are genuinely  distant from the training data set. We have also extended the robust monitor concept to a generic setting where each neuron can be monitored using an activation pattern expressed using more than one bit.  With a particular focus on operation monitoring, this work well complements existing results that also apply formal methods to improve the quality of DNNs. 

As a preliminary evaluation, we implemented the proposed technology into a lab setting, where we engineered a DNN to generate visual waypoints from images and deployed the monitor to detect abnormal situations. For the optimal case with $0.62\%$ of false positives (i.e., the vehicle is inside ODD but warning is raised), the monitor can detect (with a substantially higher rate) our additionally created scenarios that are not included in the training data set, such as dark conditions, construction site, or ice on the track (illustrated in Figure~\ref{fig.experiment}). By changing to robust monitors, the false positive rate is further reduced to $0.125\%$ (i.e., $80\%$ reduction) while the detection rate of ODD departures remains roughly the same. 
However, we also discovered that some monitors, although demonstrating~$0\%$ false positive, are inefficient in that only a few warnings are raised. We expect to conduct further studies to understand how to train networks with better monitorability and to evaluate the technique in real-world scenarios such as multi-view 3D reconstruction~\cite{chen2017multi,yang2018pixor}. This helps to understand the full potential and limitations of the approach. We also plan to integrate the correctness claims into recently developed results in logical safety argumentation~\cite{DBLP:conf/safecomp/BeyeneS20,DBLP:conf/safecomp/ZhaoBSRF0020}.


\bibliographystyle{IEEEtran}


\end{document}